%% file: main.tex
\pdfoutput=1

\documentclass[11pt]{article}
\def\shownotes{1}
\usepackage[final]{acl}
\usepackage{natbib}
\usepackage{times}
\usepackage{latexsym}

\usepackage[T1]{fontenc}

\usepackage[utf8]{inputenc}

\usepackage{microtype}
\usepackage{inconsolata}
\usepackage{macro}
\usepackage{graphicx}
\usepackage{amsmath}
\usepackage{times}
\usepackage{latexsym}
\usepackage{makecell}
\usepackage{balance}
\usepackage{fullpage}
\usepackage[T1]{fontenc}

\usepackage[utf8]{inputenc}

\usepackage{microtype}

\usepackage{inconsolata}

\usepackage{graphicx}
\usepackage{balance}
\usepackage{caption}
\usepackage{subcaption}
\usepackage{booktabs}
\usepackage{algorithm,algorithmicx,algpseudocode}
\usepackage{enumitem}
\usepackage{multirow}
\makeatletter
\def\thanksnosymbol#1{\protected@xdef\@thanks{\@thanks
        \protect\footnotetext{#1}}}
\makeatother

\usepackage{hyperref}

\title{Linear-Time Demonstration Selection for In-Context Learning via Gradient Estimation}

\author{
Ziniu Zhang\textsuperscript{*}\thanksnosymbol{\textsuperscript{*}Equal Contribution. Email correspondence can be directed to all authors at {\{zhang.zini, zhang.zhens, li.dongyu, ho.zhang\}@northeastern.edu}, {wangluxy@umich.edu}, and jdy@ece.neu.edu.}\\
Northeastern University\\\And
Zhenshuo Zhang\textsuperscript{*}\\
Northeastern University\\\And
Dongyue Li\\
Northeastern University\\\AND
Lu Wang\\
University of Michigan\\\And
Jennifer Dy\\
Northeastern University\\\And
Hongyang R. Zhang\\
Northeastern University
}

\begin{document}
\maketitle

\input{intro}
\input{content}

\bibliography{ref}
\clearpage
\appendix
\input{appendix}

\end{document}

%% file: intro.tex
\begin{abstract}
    This paper introduces an algorithm to select demonstration examples for in-context learning of a query set. Given a set of $n$ examples, how can we quickly select $k$ out of $n$ to best serve as the conditioning for downstream inference? This problem has broad applications in prompt tuning and chain-of-thought reasoning. Since model weights remain fixed during in-context learning, previous work has sought to design methods based on the similarity of token embeddings. This work proposes a new approach based on gradients of the output taken in the input embedding space. Our approach estimates model outputs through a first-order approximation using the gradients. Then, we apply this estimation to multiple randomly sampled subsets. Finally, we aggregate the sampled subset outcomes to form an influence score for each demonstration, and select $k$ most relevant examples. This procedure only requires pre-computing model outputs and gradients once, resulting in a linear-time algorithm relative to model and training set sizes. Extensive experiments across various models and datasets validate the efficiency of our approach. We show that the gradient estimation procedure yields approximations of full inference with less than ${1}\%$ error across six datasets. This allows us to scale up subset selection that would otherwise run full inference by up to ${37.7}\times$ on models with up to $34$ billion parameters, and outperform existing selection methods based on input embeddings by ${11}\%$ on average.
\end{abstract}

\section{Introduction}

Prompt-based learning has emerged as a paradigm for solving and generating natural language tasks using large language models (LLMs).
In general, prompt tuning helps adjust input prompts for a pre-trained LLM \cite{liu2023pre,wang2022learning}, allowing models to adapt to new tasks by conditioning on carefully chosen prompts at inference time, rather than undergoing full fine-tuning.
The capability of in-context learning (ICL) has been discovered in GPT-3 models, where LLMs learn to make predictions from a few examples \cite{brown2020language}, and can be theoretically fleshed out in simple function classes \cite{garg2022can}.
In practice, in-context learning is sensitive to the examples in the prompt \cite{min2022rethinking}. Small changes in the demonstration examples can lead to different outcomes (see, e.g., \citet{albalaksurvey} for a recent survey).
In this paper, we study the problem of selecting a subset of demonstration examples for in-context learning of a query set, focusing on the \emph{efficiency} of the selection procedure.

\begin{figure*}[t!]
    \centering
    \includegraphics[width=0.9995\textwidth]{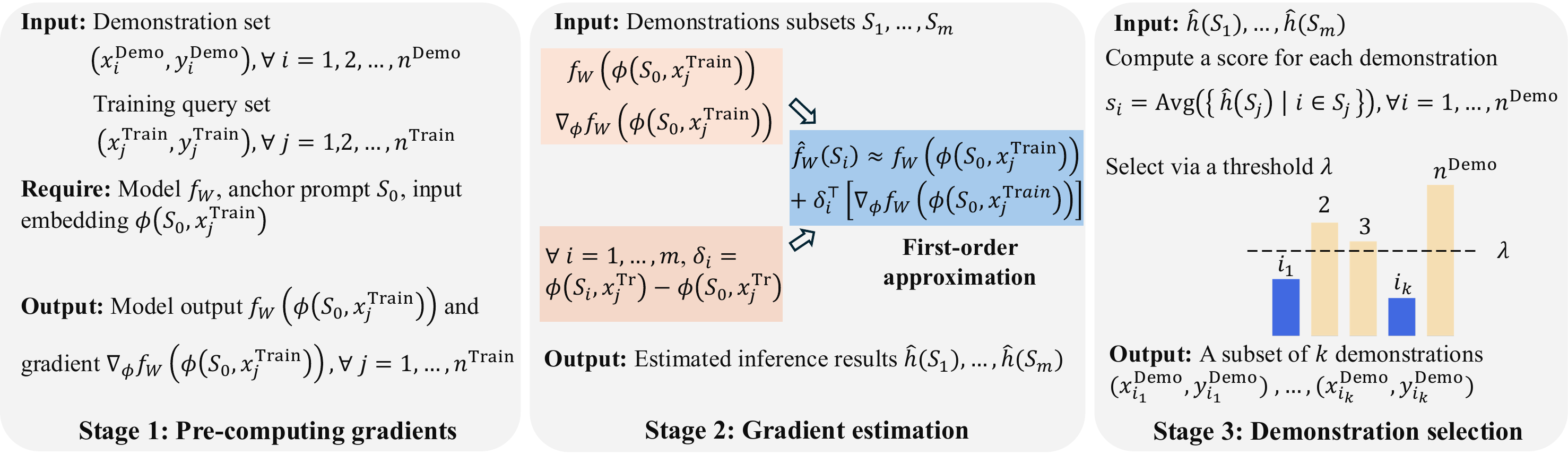}
    \caption{Given a set of demonstrations, we design a linear-time demonstration selection algorithm to construct prompts for in-context learning of a query set.
    \underline{Stage 1}: First, pre-compute functional outputs and gradients (with respect to the embedding vector) on the entire training set.
    \underline{Stage 2}: Second, apply a first-order approximation to estimate the model outputs on a list of $m$ random subsets $S_1, S_2, \dots, S_m$. This approximation is computed based on the model outputs and the gradients computed during Stage 1.
    Let $\hat{h}(S_1), \hat{h}(S_2), \ldots, \hat{h}(S_m)$ denote the estimated results, corresponding to the loss values of evaluating $f_W$ with each subset as the prompt conditioning.
    \underline{Stage 3}: Third, compute an influence score $s_i$ for each demonstration example based on the estimated $\hat h$, for $i = 1, 2, \dots, n^{\text{Demo}}$.
    Specifically, $s_i$ can be thought of as the importance score for the $i$-th demonstration example applied to the query set.
    Then, select a subset of $k$ out of $n$ demonstrations via a threshold $\lambda$ on the scores.}\label{fig_pipeline}
\end{figure*}

Demonstration selection has many applications. One example is chain-of-thought prompting, where models are guided by a few reasoning demonstrations before solving new problems. %
Another scenario is long-context learning, where LLMs process hundreds of demonstrations in a single prompt. As performance gains can diminish with increasing context length \cite{bertsch2024context}, careful selection is critical.
Finally, in-context learning is related to earlier literature on language inference \cite{roth2004linear}, but the mechanism of retrieving information from LLMs remains poorly understood \cite{garg2022can,he2022rethinking}.

Existing works have sought to leverage input embeddings to select the most relevant demonstrations for in-context learning. For instance, one could select the most similar demonstrations based on their input embeddings \cite{liu2022makes}. While this similarity-based selection can identify similar examples to the query, it overlooks the model output conditioned on the demonstration labels \cite{peng2024revisiting}. Moreover, it treats each example independently, even though in-context learning performance can depend on the interaction between multiple prompt examples. To identify such subset combinations, one might consider subset selection methods that evaluate model losses directly. For $n$ demonstrations, evaluating all $\binom{n}{k}$ subsets is infeasible. Stepwise selection methods, such as forward selection, reduce the cost to $O(kn)$ by iteratively adding examples that yield the lowest loss; however, this remains costly for large $n$. Another option is random ensemble selection \cite{li2023boosting,li2023identification}, which averages model losses over $O(n \log n)$ randomly sampled subsets, again incurring high inference cost. In summary, existing methods rely solely on input embedding similarities or suffer from high computation cost.

This work proposes a scalable prompt selection method that utilizes the gradients of the model output with respect to the input embeddings to efficiently estimate model losses.
Gradients have been used as features for computing model influence functions for data attribution \cite{li2024scalable2,li2024scalable}.
However, existing methods that use gradients to select data involve modifying model parameters \cite{li2025efficient}, rendering them unsuitable for ICL, where weights do not change.
Let $f_W$ denote the output of an LLM.
Let $\phi(S, x)$ denote the embedding of a prompt $S$ followed by a query input $x$.
Our main observation is to use gradients of $f_W(\phi(S, x))$ taken with respect to $\phi(S, x)$, and \emph{apply a first-order Taylor expansion to $f_W$ in the input embedding space}.
In particular, we estimate the model output for each training set query using a gradient-based estimation, which we also empirically verify to hold with an error of less than $1\%$ across seven LLMs (see Table \ref{table_motivation_result}).
With this estimation, we then sample $m$ random subsets from the demonstration examples, and estimate their loss values \emph{without actually running model inference at all}.
Finally, we aggregate the loss values from random subsets to form an influence score for every demonstration, and select $k$ examples with the most relevant scores.
Importantly, this procedure only requires computing the output and the gradient of every candidate example in the embedding space once at $f_W$, leading to a computation time of $3T$, where $T$ is the runtime for one forward pass on the training set, plus $O(n\log n)$ during inference, where $n$ is the number of demonstration examples.
In particular, $T$ is much larger than the runtime for computing the gradients of the demonstration examples, resulting in a linear running time.
See Figure \ref{fig_pipeline} for an illustration of our approach.

We extensively evaluate the efficiency and accuracy of our approach on LLMs across three sentiment classification datasets and three math reasoning tasks. 
Our approach achieves up to $\textbf{37.7}\times$ speed-up compared to conducting subset selection with full-model inference, such as forward selection and random ensemble selection, while yielding an approximation error of less than $\textbf{1\%}$.
For in-context learning evaluations on the six datasets, our approach outperforms the strongest baseline of top-$k$ followed by loss-based selection \cite{peng2024revisiting} by $\textbf{11\%}$ on average, while using $49\%$ less computation cost.
In long context evaluations where $k = 150$, our approach matches the performance of existing baselines with $\textbf{30}\times$ shorter context length.

In summary, the contributions of this paper include:
i) Introducing a gradient estimation procedure to scale up inference and empirically validate the accuracy of first-order approximations in input embedding spaces.
ii) Designing a linear-time demonstration selection algorithm based on random ensemble aggregation. See Table \ref{tab_runtime_comparison} for a detailed runtime comparison.
iii) Extensive experiments validating the efficiency of gradient estimation for model inference and demonstration selection.
We provide the code to replicate these findings at \href{https://github.com/VirtuosoResearch/ICL-GradSel}{https://github.com/VirtuosoResearch/ICL-GradSel}.

\begin{table}[t!]
\centering
\caption{Runtime comparison between our algorithms and standard subset selection methods for demonstration selection.
Here, $T$ represents the time cost of running one forward pass on a model with the training set, $k$ denotes the number of selected examples, and $n$ represents the number of demonstrations. Our approach computes model outputs with one forward pass of time $T$ and input gradients with time $2T$, plus an overhead cost of $O(n \log n)$, which is negligible relative to $T$.}\label{tab_runtime_comparison} %
{\begin{tabular}{@{}l c@{}}
\toprule
\textbf{Approach} & \textbf{Runtime} \\
\midrule
Forward Stepwise Selection & $n k T$ \\
Random Ensemble Selection & $O\big(n (\log n) T \big)$ \\
Ours (Algorithm \ref{alg_forward_selection}) & $\approx 3kT$ \\
Ours (Algorithm \ref{alg_random_ensemble}) & $\approx \bm{3T}$ \\
\bottomrule
\end{tabular}}
\end{table}

%% file: content.tex
\section{Preliminaries}\label{sec_problem_setup}

We consider the problem of solving a downstream task through in-context learning ({ICL}).
During ICL, the prompt given to a language model appears in the form of $(x_1, y_1, \dots, x_k, y_k, x_{\text{query}})$.
The model is asked to provide an answer to the query $x_{\text{query}}$, based on the preceding $k$ demonstration examples.
In general, suppose we have access to a set of $n^{\text{Demo}}$ demonstration examples, denoted by $\cS^{\text{Demo}}$, including $(x^{\text{Demo}}_i, y^{\text{Demo}}_i)$, for $i = 1, \dots, n^{\text{Demo}}$.
At training time, we have a set of $n^{\text{Train}}$ training examples, denoted as $\cS^{\text{Train}} = \set{(x^{\text{Train}}_i, y^{\text{Train}}_i)}_{i=1}^{n^{\text{Train}}}$.
Since the demonstration examples are not always helpful at training time \cite{min2022rethinking}, an important consideration is to select a subset $S$ out of $S^{\text{Demo}}$, and use this subset as the prompt conditioning instead of the entire set.
More formally, let $f_W$ denote a pretrained language model and let $\ell$ denote a loss function.
The input to $f_W$ includes a prompt conditioning sequence $S$, followed by a query example $x_{\text{query}}$ from the training set.
In practice, an embedding function, denoted as $\phi$, is used to encode the prompt sequence along with the query example to the language model, leading to the training objective of: 
\[ h(S) = \frac 1 {n^{\text{Train}}} \sum_{i=1}^{n^{\text{Train}}} \ell\big(f_W(\phi(S, x_i)), y_i\big). \]
Let $p$ denote the embedding dimension.
Our goal is to select a subset $S\subseteq\cS^{\text{Demo}}$ that minimizes the above objective.
After selecting a set of demonstrations, we evaluate the test performance on a separate test set.

In general, the number of all possible subsets of demonstration examples is exponential.
Enumerating through all possible subsets thus requires performing inference through $f_W$ on every subset.
How can we scale up the inference procedure for faster demonstration selection?
Suppose there is a list of subsets $S_1, \dots, S_m \subseteq \cS^{\text{Demo}}$, how fast can we estimate $h(S_1), \dots, h(S_m)$?
To give several examples of subset selection, suppose we use greedy selection. The subset selection procedure would start from the empty set, then include all the singleton sets, then a single demonstration combined with another demonstration, and so on.
Another example would be to take random subsets out of $\cS^{\text{Demo}}$, and then use the outcome from the random subsets to form an influence score for every demonstration example \cite{li2024scalable,li2025efficient}.
Even if we were to run inference to compute $h(S_1), \dots, h(S_m)$, it would still be slow if the size of $f_W$ is very large.
Therefore, we ask the following question: \emph{Is it possible to estimate the model outputs accurately without having to run through full inference on all the subsets?}

\section{Scaling Up Inference via Estimation}\label{sec_approach}

In this section, we introduce a new algorithm for estimating model inference outcomes without requiring repeated inference on multiple subsets.
The key idea is to estimate $f_W$ based on a first-order approximation property.
Consider the model output of a prompt subset $S$ on an input $x$, denoted as $f_W(\phi(S, x))$.
Given an anchor prompt $S_0$, the first-order approximation of $f_W$ around the embedding vector $\phi(S_0, x)$ is given by:
\begin{align}
    & f_W(\phi(S, x)) = f_W(\phi(S_0, x)) + \label{eq_fo}\\
    & \biginner{\nabla_{\phi}f_W(\phi(S_0, x))}{ \phi(S, x) - \phi(S_0, x)} + \epsilon_{S, x},\nonumber
\end{align}
where $\nabla_\phi$ denotes the gradient of $f_W$ with respect to the input embedding $\phi$ and $\epsilon_{S, x}$ denotes the approximation error. As a remark, we add padding tokens to the right of the prompts to ensure that all the prompts have the same length.

\smallskip
\noindent\textbf{First-order approximation is accurate for ICL.}
We find that $\epsilon_{S, x}$ remains small for a wide range of LLMs and datasets.
We evaluate $\epsilon_{S, x}$ across seven LLMs with $1$ billion up to $34$ billion parameters, evaluated on the SST-2 dataset from the GLUE benchmark \cite{wangglue}.
We randomly sample a reference prompt of size $k = 50$ and evaluate equation \eqref{eq_fo} for other randomly sampled prompts with $k = 50$.
We compute ${\epsilon}_{S, x}$ and report the relative error $\big(\frac{\epsilon_S} {f_W(\phi(S, x))}\big)^2$, averaged over the training set $\cS^{\text{Train}}$. Here, $f_W$ refers to the model output; For binary classification, it refers to the value of the logit function, which is a scalar. For multi-class classification, one would instead compute the output and the approximation error at each label position.
We normalize the error with respect to the norm of the embedding vector, i.e., $\frac{\bignorm{\phi(S, x) - \phi(S_0, x)}} {\bignorm{\phi(S_0, x)}}$, as shown in the leftmost column of Table \ref{table_motivation_result}.
Similar results are observed on other datasets and are described in detail in Table \ref{table_approx_first_order}, Appendix \ref{app_omitted_approximation_error}.

The results are shown in Table \ref{table_motivation_result}.
We find that the estimation error remains less than $\textbf{1}\%$.
Additionally, we find that the model with 34 billion parameters yields the smallest error, achieving less than $\textbf{0.3}\%$.
These results suggest that equation \eqref{eq_fo} delivers an accurate approximation for ICL.
\begin{table}[h!]
\centering
\caption{Relative approximation error of $\epsilon_{S, x}$, tested on several datasets with language models of up to $34$ billion parameters.}\label{table_motivation_result}
\resizebox{1.00\columnwidth}{!}
{%
\begin{tabular}{@{}lccc@{}}
\toprule
{Distance} & {DeepSeek-7B} & {Llama-13B} & {CodeLlama-34B}\\ 
\midrule
$15\%-20\%$    & $0.21_{\pm0.04} \%$ & $0.06_{\pm0.00} \%$ & $0.06_{\pm0.01} \%$ \\
$20\%-25\%$    & $0.30_{\pm0.05} \%$ & $0.08_{\pm0.02} \%$ & $0.06_{\pm0.01} \%$ \\
$25\%-30\%$    & $0.51_{\pm0.04} \%$ & $0.09_{\pm0.01} \%$ & $0.08_{\pm0.02} \%$ \\
$30\%-35\%$    & $0.53_{\pm0.07} \%$ & $0.28_{\pm0.03} \%$ & $0.13_{\pm0.03} \%$ \\
$35\%-40\%$    & $1.40_{\pm0.20} \%$ & $0.37_{\pm0.03} \%$ & $0.30_{\pm0.03} \%$ \\
\bottomrule
    \end{tabular}}
\end{table}

Next, we describe the design of the estimation algorithm.
Given a reference prompt $S_0$, we apply equation \eqref{eq_fo} to every $S_i$, for $i = 1, 2, \dots, m$.
In order to enable this approximation, we pre-compute the outputs at the reference prompt on the training set, as well as the gradients at the reference prompt, for every training example in $\cS^{\text{Train}}$, taken with respect to the embedding function $\phi$.
In the case that $k$ is particularly large, so that the dimension of $\phi$ becomes very high, we can use random projection to reduce the dimension of the gradients down to a few hundred, which provably preserves the Euclidean geometry between the gradient vectors.
By the Johnson-Lindenstrauss lemma, random projection can provably preserve the Euclidean distance between the gradient vectors after the projection \citep{johnson1984extensions}.
See the complete statement in Appendix \ref{app_JL}.
This estimation procedure incurs less than $30\%$ increase in memory usage compared to standard model inference.
The overall procedure is described in Algorithm \ref{alg_gradient_estimation}.

\begin{algorithm}[t!]
\raggedright
\caption{In-Context Learning via Gradient Estimation (\estimation{})}\label{alg_gradient_estimation}
\textbf{Input:} An anchor set $S_0\subseteq\cS^{\text{Demo}}$; $m$ subsets $S_1, S_2, \dots, S_m\subseteq \cS^{\text{Demo}}$\\ 
\textbf{Output:} Estimated inference results for $\set{S_i}_{i=1}^m$\\
{/*\qquad\qquad\quad~{\itshape \underline{Stage 1: Pre-compute}} \hfill */}
\begin{algorithmic}[1]
    \State $P \subseteq\real^{p\times d} \leftarrow$ Isotropic Gaussian matrix whose entries are drawn from $\cN(0, 1)$ independently
    \For{$i = 1, 2, \dots, n^{\text{Train}}$}
        \State $f_W(\phi(S_0, x^{\text{Train}}_i)) \leftarrow$ Model output
        \State $\nabla_\phi f_W(\phi(S_0, x^{\text{Train}}_i)) \leftarrow$ Gradient over the embedding space $\phi$
        \State $\tilde{g}_i \leftarrow P^\top \nabla_\phi f_W(\phi(S_0, x^{\text{Train}}_i))$ %
    \EndFor
\end{algorithmic}
{/*\qquad\qquad\quad~{\itshape \underline{Stage 2: Inference}} \hfill */}
\begin{algorithmic}[1]
    \For{$j = 1, 2, \dots, m$}
        \For{$i = 1, 2, \dots, n^{\text{Train}}$}
            \State $\delta_{i, j} \leftarrow \phi(S_j, x^{\text{Train}}_i) - \phi(S_0, x^{\text{Train}}_i)$
            \State $\tilde{\delta}_{i, j} \leftarrow P^\top \delta_{i, j}$
            \State $\hat{f}_{i,j} \leftarrow f_W(\phi(S_0, x^{\text{Train}}_i)) + \inner{\tilde{g}_i}{\tilde{\delta}_{i, j}}$
            \State $\hat{h}(S_j) \leftarrow \frac 1 {n^{\text{Train}}} \sum_{i=1}^{n^{\text{Train}}} \ell(\hat f_{i, j}, y^{\text{Train}}_i)$ 
        \EndFor
    \EndFor
    \State \textbf{return} $\hat{h}(S_1), \hat{h}(S_2), \dots, \hat{h}(S_m)$
\end{algorithmic}
\end{algorithm}

\smallskip
\noindent\textbf{A case study of linear functions.}
We illustrate the above approximation with an example of learning linear functions with transformer models \cite{garg2022can}.
We use this setting to evaluate the accuracy of equation \eqref{eq_fo} for ICL of linear functions.
Suppose that both the in-context example and query example are generated as $y_i = \inner{\beta^{(j)}}{x_i} + \epsilon_i$, where the input vector $x_i$ is randomly sampled from $\mathcal{N}(0, \id_p)$ with $p = 20$. The coefficient vector $\beta^{(j)}$ is randomly drawn from a finite set $\{\beta^{(1)}, \beta^{(2)}, \cdots, \beta^{(C)}\}$, meaning that different in-context examples may have different coefficients. We consider two settings for the noise term $\epsilon_i$: linear regression with $\epsilon_i = 0$ for all $i$, or noisy linear regression with $\epsilon_i \sim \mathcal{N}(0, 1)$.

Figure \ref{fig_lr} shows that in both settings, \estimation{} can accurately estimate the true outcomes of ICL.
In particular, as $k$ increases, the trained transformer learns the linear function correctly, and the residual sum of squares (RSS) between the estimated scores and the true scores gradually decreases to near zero.
Recall that $p = 20$. We thus plot from $k = p$ onward because this is the minimum number of samples required to solve a system of linear equations.
When $k < p$, the trained transformer is unable to fully determine the linear function. (See Figure 2, \citet{garg2022can}).

\smallskip
\noindent\textbf{Extension to nonlinear functions.}
The same approach is applicable to more complex, non-linear function classes. We extend the case study to two-layer ReLU neural networks. \estimation{} achieves an estimation error of $4\%$ and outperforms the top-$k$ and random-$k$ baselines by $34\%$.

\section{Efficient Demonstration Selection}

\begin{figure}[t!]
    \centering
    \begin{subfigure}[b]{0.24\textwidth}
        \centering
        \includegraphics[width=0.99\textwidth]{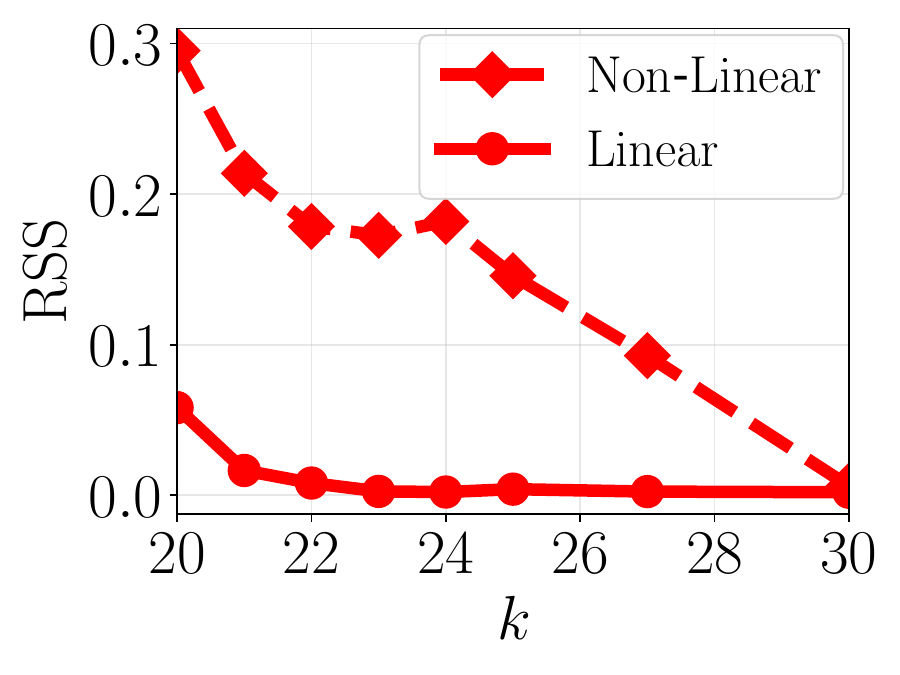}
        \caption{Approximation Error}\label{fig_lr}
    \end{subfigure}\hfill
    \begin{subfigure}[b]{0.24\textwidth}
        \centering\includegraphics[width=\textwidth]{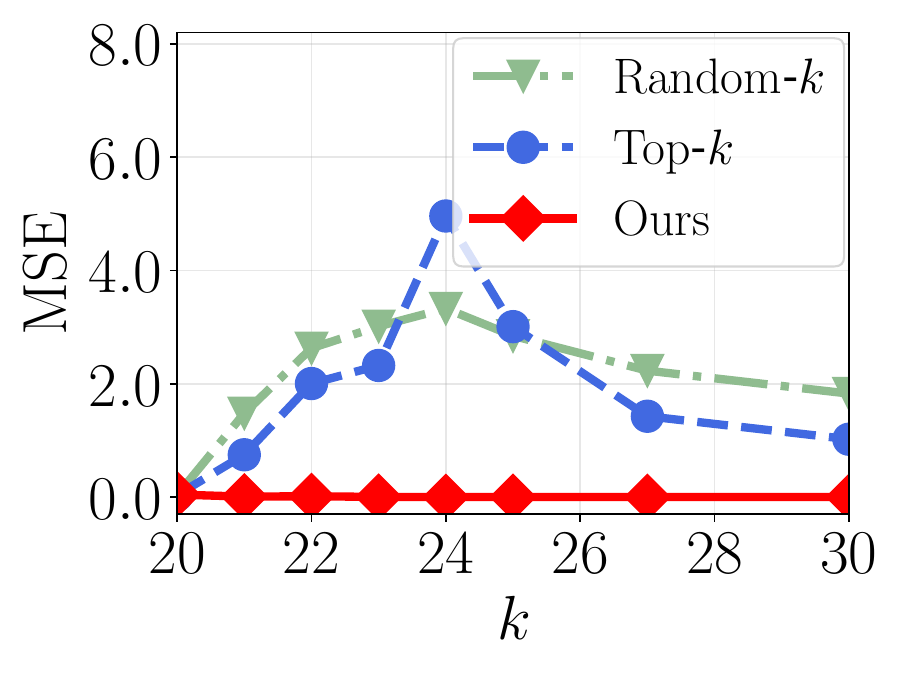}
        \caption{Query Error}\label{fig_lr_selection}
    \end{subfigure}\hfill
    \caption{An illustration of our approach for in-context learning of linear functions, as we vary the number of in-context examples $k$.
    Figure \ref{fig_lr}: \estimation{} incurs low approximation error relative to full inference, for both linear and nonlinear regression.
    Figure \ref{fig_lr_selection}: With \acronym{}, the selected demonstrations follow the same linear function $\beta$, achieving lower error than top-$k$ and random-$k$ selections.}\label{fig_incontext_results}
\end{figure}

We now describe our design of fast demonstration selection methods using a first-order approximation technique.
The first, called \acronymre{}, applies \estimation{} to the random ensemble, which is inspired by the literature on influence functions~\cite{li2023identification} and the work of task affinity grouping~\cite{fifty2021efficiently,li2025efficient}.
We draw $m$ random subsets from $\{1, 2, \dots, n\}$, each of fixed size $k$. Then, we use \estimation{} to estimate $\hat{h}(S)$ for each of the subsets and compute a score for a demonstration $\theta_i$ as the average of $\hat{h}(S)$ over subsets $S$ that include $s_i$. Then, we choose a subset of size $k$ with the lowest scores.

To estimate each subset’s inference loss, we first sample $\alpha$ anchor subsets of size $k$, then apply \estimation{} to estimate $\hat{h}(S)$ from the anchor subset for each subset $S_j$, for $j=1, 2, \ldots, m$. Considering the robustness of this random selection process, we observe that even when the average distance is around $40\%$, the approximation error remains below $1.5\%$, as shown in Table~\ref{table_motivation_result}. To obtain better results, we use the average estimates from the $\alpha$ anchor subsets. Our ablation studies confirm that $\alpha$ is typically a small constant within $10$. On the SST-2 dataset, setting $\alpha$ to over 5 is sufficient for convergence.
The complete procedure is described in Algorithm \ref{alg_random_ensemble}.

Next, we illustrate the algorithm in the case of linear functions.
We sample $3,000$ data points from three linear function of different $\beta^{(1)}, \beta^{(2)}, \beta^{(3)} $ in $\real^{20}$ as the demonstration set, and sample $100$ data points from the linear function of $\beta^{(1)}$ as the training query set.
Then, we apply our approach to select a subset of $k$ from the demonstration set to construct the prompt for ICL.
We compare \acronymre{} with top-$k$ selection, which chooses demonstrations based on input embedding similarity to the training query set, and random-$k$, which randomly samples $k$ demonstrations.
The results in Figure \ref{fig_lr_selection} show that our approach finds in-context examples with the same $\beta^{(1)}$ out of the demonstration set, whereas top-$k$ and random-$k$ selections are sensitive to different values of $k$.
To demonstrate the robustness of the random ensemble method, we calculate the scores of good examples (those following the same linear model) versus those drawn from different linear models, while varying the number of subsets.
Our results indicate that the scores for in-distribution examples are consistently lower than those for out-of-distribution examples (See Appendix~\ref{app_ablation}).
\begin{algorithm}[t!]
\raggedright
\caption{In-Context Learning via Gradient-based Random Ensemble (\acronymre{})}\label{alg_random_ensemble}
\textbf{Input:} Query set $\cS^{\text{Train}}$; Demonstration set $\cS^{\text{Demo}}$\\
\textbf{Require:} Pretrained model $f_W$; Projection dimension $d$; Number of subsets $m$; Subset size $k$; Number of anchor prompts $\alpha$\\
\textbf{Output:} Selected subset $S \subseteq \cS^{\text{Demo}}$ of size $k$
\begin{algorithmic}[1]
    \State $S_1, S_2, \dots, S_m \leftarrow$ Sample $m$ random subsets each of size $k$ from $\cS^{\text{Demo}}$
    \State ${A^{(1)}, A^{(2)}, \dots, A^{(\alpha)}} \leftarrow$ Randomly choose $\alpha$ anchor subsets from $S_1, S_2, \dots, S_m$
    \State $\hat h^{(i)}(S_j) \leftarrow$ {\estimation{$({A^{(i)}}, {S_j})$}}, for any $1\le i\le \alpha$ and $1\le j\le m$
    \For{$q = 1, 2, \dots, n^{\text{Demo}}$}
        \State $s_q \leftarrow \frac 1 {\alpha} \sum_{i=1}^{\alpha}$ Avg$\big(\bigset{\hat h^{(i)}(S_j) \mid \forall (x_q^{\text{Demo}}, y_q^{\text{Demo}}) \in S_j}\big)$
    \EndFor
    \State $S_k\subseteq\cS^{\text{Demo}} \leftarrow$ $k$ demonstrations with the lowest-$k$ scores ranked in $s_1, s_2, \dots, s_{n^{\text{Demo}}}$
    \State \Return $S_k$
\end{algorithmic}
\end{algorithm}

Besides random ensembles, our estimation approach can also be instantiated to accelerate other subset selection methods.
One, \acronymfs{}, implements the classical forward selection procedure.
At each step, we use \estimation{} to all the subsets that are encountered during the search step.
Then, based on the estimated results, we select a demonstration example that leads to the smallest (estimated) loss and add that to the chosen subset.
This procedure continues until we have selected $k$ demonstrations.
The other, named \acronymcone{}, uses \estimation{} to accelerate a cross-entropy loss-based selection.
The detailed procedure of these two algorithms is deferred to Appendix \ref{app_omitted_approximation_error}.

\section{Experiments}\label{sec_exp}

We now evaluate \estimation{} and \acronymre{} across various LLMs and datasets.
The evaluation focuses on answering the following questions.
First, how accurate are gradient estimations as compared to full model inference results?
And how much computation cost does it save compared to running an otherwise accurate subset selection procedure with full model inference?
Second, in addition to the estimated results, how effective is the selected demonstration set for in-context learning?

Our experiments show that, across a wide range of LLMs (such as Llama, OPT, DeepSeek, and Qwen), \estimation{} can estimate model inference outcomes within $\textbf{1}\%$ error, for LLMs ranging from $1$ billion up to $34$ billion parameters.
Crucially, this estimation allows us to reduce the computation cost of subset selection by up to $\textbf{37.7}\times$.
When applied to in-context learning across six datasets that include sentence classification and math reasoning tasks, \acronymre{} improves the test $F_1$ score by $\textbf{11}\%$ on average.
For long-context settings where $k = 150$, our selection matches the performance of top-$k$ and random-$k$ selection with $\textbf{30}\times$ less context length.

\subsection{Experimental Setup}

\textit{Datasets and models.} 
We evaluate our algorithm on sentiment classification and reasoning tasks. For sentiment classification, we use the Poem Sentiment, SST-2, and CR datasets. For reasoning, we include edge existence from the GraphQA benchmark, modular addition, and coin flip tasks.
We use pretrained LLMs including Llama, DeepSeek, and Qwen, covering model sizes ranging from $1$ to $34$ billion.
The sources of the datasets and models are described in Appendix \ref{appendix:datasets_models}.

\smallskip\noindent\textit{Baselines.} We first compare our methods to subset selection methods, including forward selection and random ensemble.
Second, we consider existing baselines based on different measures to rank the demonstrations. These include selection based on probabilistic relevance rankings (BM25), embedding similarities (top-$k$) \cite{liu2022makes}, conditional entropy (top-$k$ + CE) \cite{peng2024revisiting}, dynamic uncertainty ranking (UR) \cite{yu2025dynamic}, and BRIDGE~\cite{wan2025from}.
We also report the results from random selection with a fixed $k$.

\smallskip\noindent\textit{Implementations.} For \acronymfs{}, we apply the estimation method at each iteration. We use the prompt selected in the previous iteration, adding a new random demonstration as the anchor.
We vary the step of estimation $t$ from 1 to the length of the demonstrations.
For \acronymre{}, we estimate the model inference loss on $m$ subsets of size $k$. Among them, we select $\alpha$ subsets as anchor subsets, for which we conduct full inference to estimate the remaining results. We use $m$ between $500$ to $2,000$, subset size $k$ between $3$ to $8$, and the number of anchors between $1$ to $10$. In all our methods, we project the gradients to $d = 400$.

\subsection{Results on Estimation and Efficiency}

We evaluate the accelerated version of our approach against full inference in terms of both approximation accuracy and computational cost.
For measuring approximation accuracy, we report the relative error between the estimated loss $\hat{h}(S)$ and the true loss $h(S)$ as $\frac{1}{m} \sum_{i=1}^m \frac{(h(S) - \hat{h}(S))^2} {(h(S))^2}$.
To measure computational cost, we count the total number of floating-point operations (FLOPs) performed on an Nvidia A6000 GPU card.
We report these metrics for forward selection, random ensemble, and loss-based selection (top-$k$+CE), relative to the results from using \estimation{}.

As shown in Table~\ref{table_inference_error}, on both SST-2 and Coin-Flip datasets, our accelerated version yields accurate approximations that are within $\textbf{1}\%$ error of model inference results, for LLMs with up to 34 billion parameters.
The speed-up ratio remains consistent across different LLMs for each dataset, which all use the same number of forward passes.

Next, we report the computation cost.
For DeepSeek-7B on the Coin-Flip dataset, \acronymfs{} takes $1.6$ GPU hours, achieving a $\textbf{37.1}\times$ reduction compared to forward selection with full inference.
\acronymre{} takes just $0.1$ GPU hours, offering a $\textbf{19.7}\times$ speed-up.
\acronymcone{} runs in $0.8$ GPU hours, reducing cost by $\textbf{17.3}\times$.
Again, these speed-ups are consistent across models, as they primarily depend on the reduced number of forward passes, which remains fixed per dataset.
The number of FLOPs is reported in Table~\ref{table_computation_cost} and Appendix~\ref{app_computation}.

\subsection{Results on Demonstration Selection}

\begin{figure}[t!]
    \centering
    \begin{subfigure}[b]{0.490\textwidth}
    \begin{minipage}[b]{0.49\textwidth}
        \centering
        \includegraphics[width=0.95\textwidth]{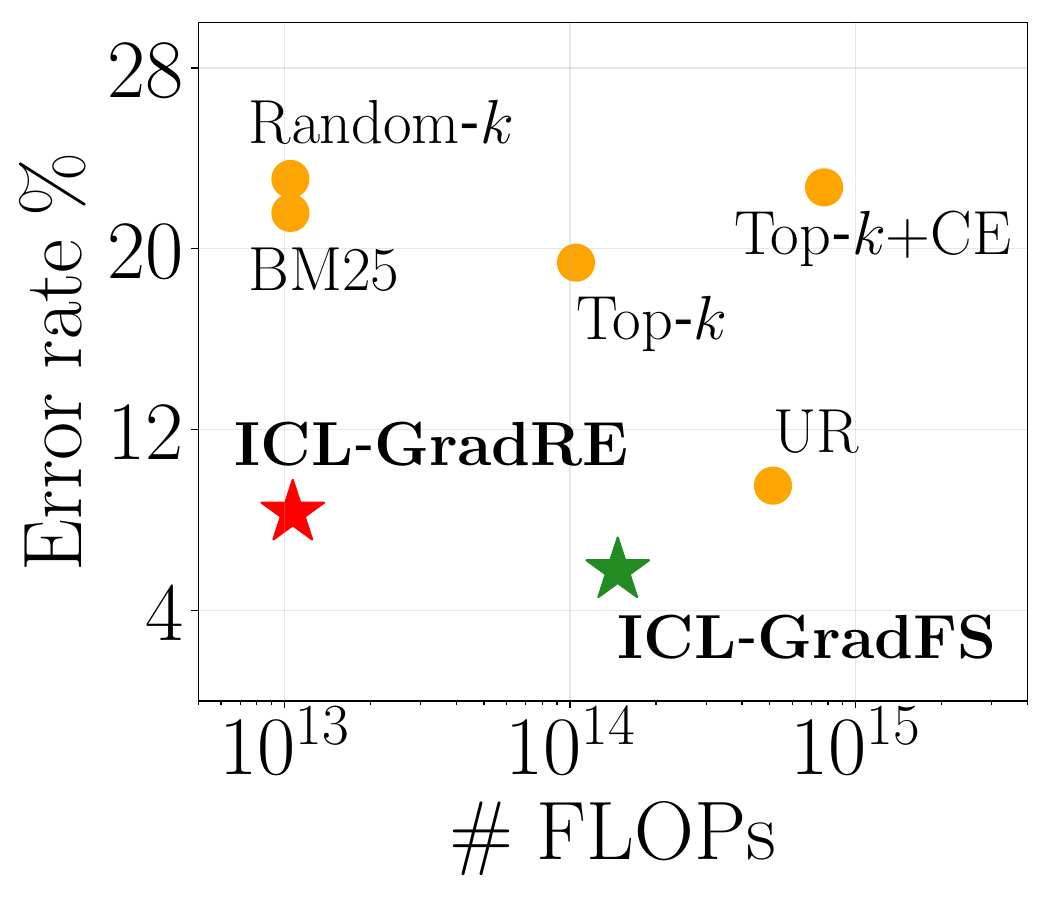}
        \caption{SST-2}
    \end{minipage}\hfill
    \begin{minipage}[b]{0.490\textwidth}
        \centering
        \includegraphics[width=0.95\textwidth]{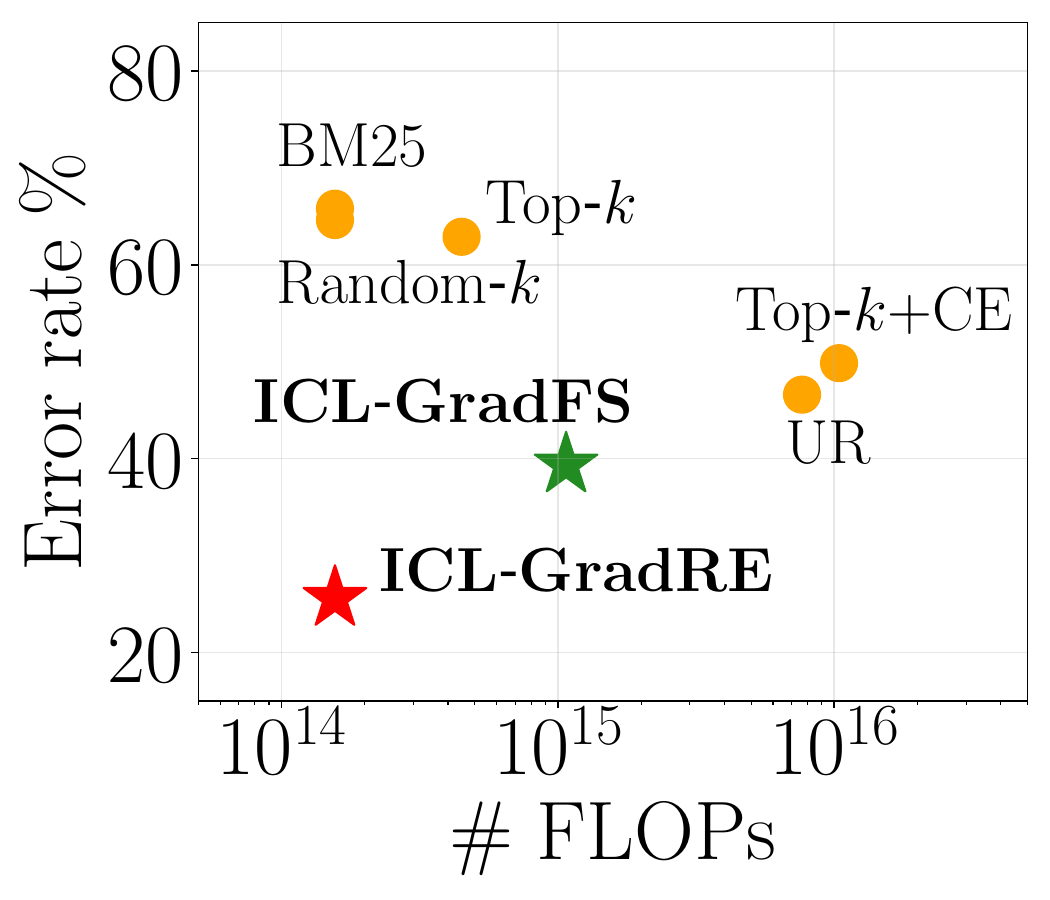}
        \caption{Coin Flip}
    \end{minipage}
    \end{subfigure}
    \caption{Trade-off between the number of FLOPs and test error rates, measured on two datasets with DeepSeek-7B models.}\label{fig_computation_performance_tradeoff}\label{tradeoff_figure}
\end{figure}

Next, we report the results of demonstration selection for in-context learning.
Table~\ref{tab_main_result} shows the comparative results across six datasets, while Figure \ref{fig_computation_performance_tradeoff} illustrates the computation vs. performance trade-off. See also Table \ref{table_computation_cost} for the number of FLOPs used to represent this figure.

First, compared to baselines that rank demonstrations by input relevance or similarity (e.g., BM25, top-$k$, and uncertainty ranking), both \acronymfs{} and \acronymre{} can outperform them by an average of $8\%$ and $11\%$, while also using $40\%$ and $49\%$ less computation, respectively.

For top-$k$+CE, which first reduces the demonstration set down to a reduced set, and then applies cross-entropy to select the $k$ prompts, our accelerated version, \acronymcone{}, achieves comparable performance, while reducing computation by $88\%$.

Second, we report the memory cost of \acronymre{}, corresponding to the results in Table~\ref{tab_main_result}. For SST-2, CR, and Coin Flip, the memory cost is $11$ GB. For Poem Sentiment, Edge Existence, and Modular Addition, the memory cost is $13$ GB.

\begin{table}[t!]
\centering
\caption{Relative error between estimated and actual inference results, measured on two datasets. The speed-up rate is measured as the ratio of FLOPs between full inference and our estimation.}\label{table_inference_error}
\resizebox{1.00\columnwidth}{!}
{\small\begin{tabular}{@{}lcc|c@{}}
\toprule
    Coin-Flip &  {{DeepSeek-7B}} & {{CodeLlama-34B}} & {Speedup} \\ \midrule
    \acronymfs{}  & $0.17_{\pm0.0} \%$  & $0.52_{\pm0.0} \%$ & $\mathbf{37.7\times}$ \\
    \acronymre{}   & $0.08_{\pm0.1} \%$  & $0.07_{\pm0.0} \%$ & $\mathbf{19.7\times}$ \\
    \acronymcone{} & $0.58_{\pm0.0} \%$  & $0.46_{\pm0.0} \%$ & $\mathbf{17.3\times}$ \\ \midrule
    SST-2 & {{DeepSeek-7B}} & {{CodeLlama-34B}} & {Speedup} \\ \midrule
    \acronymfs{}   & $ 0.20_{\pm0.1} \%$  & $ 0.43_{\pm0.1} \%$  & $\mathbf{19.0\times}$ \\
    \acronymre{}   & $ 0.18_{\pm0.0} \%$  & $ 0.36_{\pm0.0} \%$ & $\mathbf{10.8\times}$ \\
    \acronymcone{} & $ 0.80_{\pm0.0} \%$  & $ 0.79_{\pm0.0} \%$ & $\mathbf{6.7\times}$ \\
    \bottomrule
\end{tabular}}
\end{table}

\begin{table*}[t!]
\centering
\caption{We report the test $F_1$ score ($\%$) of in-context learning using the DeepSeek-7B model across six datasets. We compare our approach with existing demonstration selection methods based on various ranking criteria.
We vary the number of in-context examples $k$ from 3 to 8 and report the best performance result for each baseline. We run each experiment with three random seeds to report the standard deviations.}\label{tab_main_result}
\resizebox{2.05\columnwidth}{!}
{\small\begin{tabular}{@{}lcccccc@{}}
\toprule
{Dataset} & {Poem Sentiment} & {SST-2} & {CR} & {Edge Existence} & {Modular Addition} & {Coin Flip}\\
{Category} & \makecell{Classification} & \makecell{Classification} & \makecell{Classification} & \makecell{Graphs} & \makecell{Math} & \makecell{Math} \\
\midrule
Random-$k$      & $55.3_{\pm 7.8}$ & $76.9_{\pm 1.0}$ & $78.5_{\pm 2.8}$ & $85.5_{\pm 2.2}$ & $66.9_{\pm 0.6}$ & $35.3_{\pm 2.1}$ \\
BM25                  & $26.5_{\pm 1.0}$ & $77.4_{\pm 2.7}$ & $76.0_{\pm 2.7}$ & $50.6_{\pm 1.3}$ & $38.1_{\pm 3.0}$ & $34.2_{\pm 1.1}$ \\
Top-$k$               & $26.3_{\pm 1.1}$ & $80.6_{\pm 2.1}$ & $88.8_{\pm 0.9}$ & $83.5_{\pm 0.1}$ & $52.2_{\pm 4.2}$ & $37.1_{\pm 2.5}$ \\
UR   & $36.7_{\pm 1.7}$ & $90.4_{\pm 1.5}$ & $93.4_{\pm 1.4}$ & $61.8_{\pm 1.4}$ & $48.1_{\pm 0.1}$& $53.4_{\pm 1.6}$ \\
BRIDGE & $26.8_{\pm 1.1}$ & $81.0_{\pm 1.3}$ & ${94.9}_{\pm 0.6}$ & $86.1_{\pm 2.5}$ & $61.0_{\pm 3.5}$ & $74.1_{\pm 2.3}$\\
Top-$k$ + CE & $44.0_{\pm 0.0}$ & $77.2_{\pm 3.5}$ & $90.7_{\pm 1.9}$ & $87.6_{\pm 2.3}$ & $67.5_{\pm 2.7}$ & $50.1_{\pm 0.0}$ \\
\acronymcone{}        & $43.3_{\pm 0.7}$ & $75.9_{\pm 2.0}$ & $90.5_{\pm 2.7}$ & $87.6_{\pm 2.5}$ & $64.1_{\pm 2.2}$ & $47.6_{\pm 1.5}$ \\
\midrule
\acronymfs{}          & $74.1_{\pm 0.0}$ & $\mathbf{94.2}_{\pm 0.7}$ & ${95.7}_{\pm 1.0}$ & ${87.6}_{\pm 2.8}$ & ${73.0}_{\pm 0.0}$ & ${60.6}_{\pm 0.0}$ \\
\acronymre{}          & $\mathbf{76.6}_{\pm 0.3}$ & $91.6_{\pm 1.8}$ & $\mathbf{97.0}_{\pm 1.3}$  & $\mathbf{89.6}_{\pm 0.6}$ & $\mathbf{80.1}_{\pm 1.3}$ & $\mathbf{75.2}_{\pm 3.7}$ \\
\bottomrule
\end{tabular}}
\end{table*}

\begin{figure}[t!]
    \centering
    \begin{subfigure}[b]{0.490\textwidth}
    \begin{minipage}[b]{0.49\textwidth}
        \centering
        \includegraphics[width=0.95\textwidth]{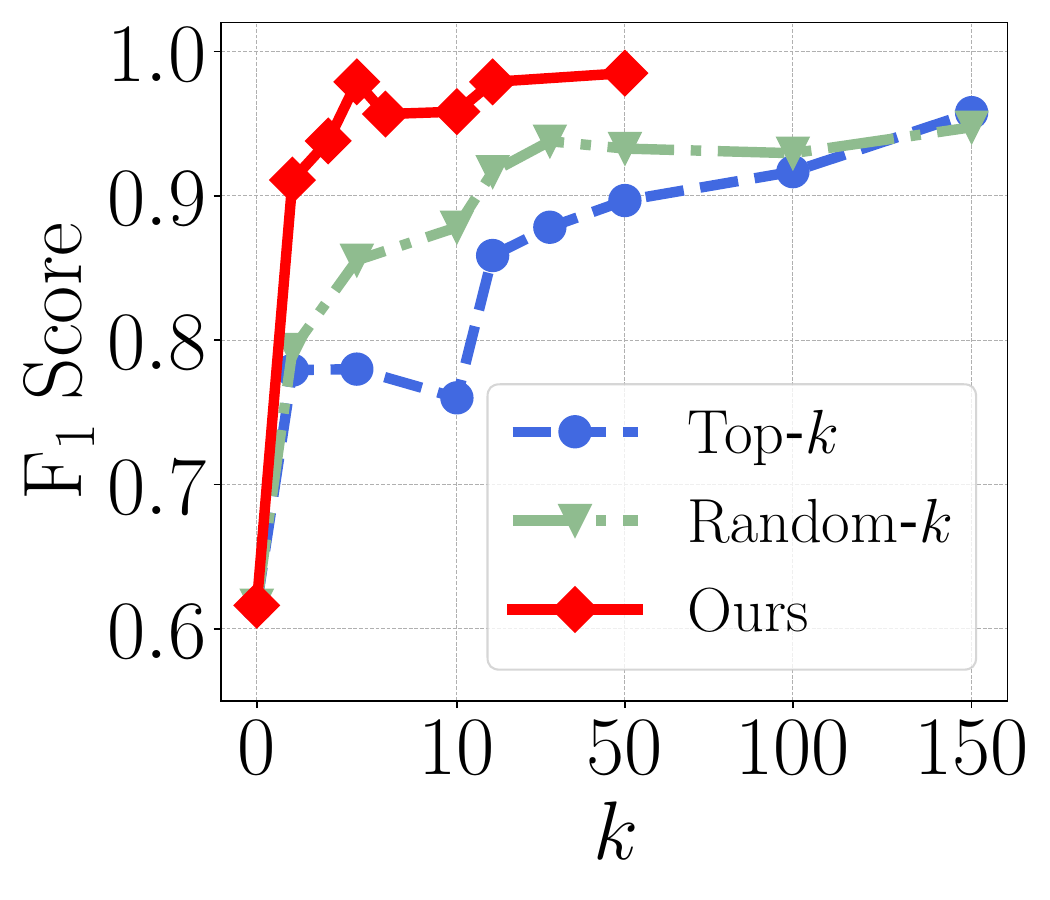}
        \caption{SST-2}
    \end{minipage}\hfill
    \begin{minipage}[b]{0.490\textwidth}
        \centering
        \includegraphics[width=0.95\textwidth]{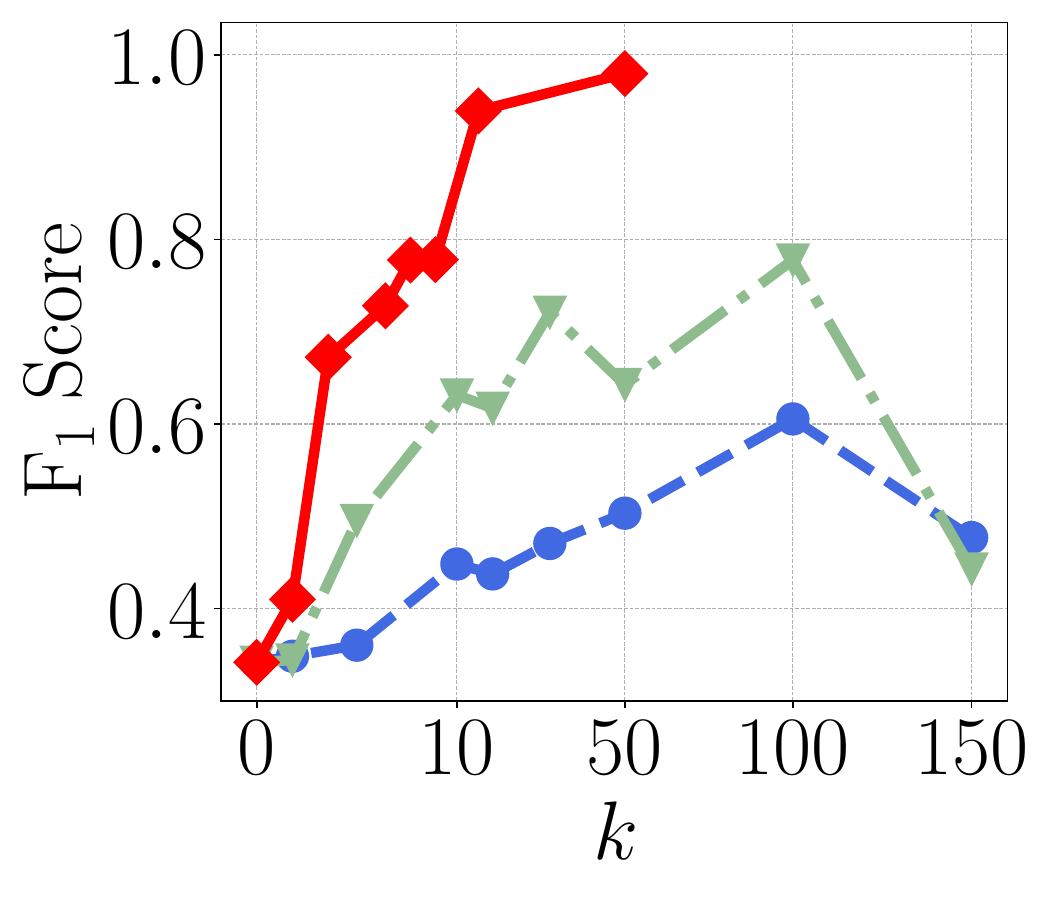}
        \caption{Coin Flip}
    \end{minipage}
    \end{subfigure}
    \caption{Comparing our approach with top-$k$ and random-$k$ by varying $k$ on two datasets using DeepSeek-7B models. Here, $k$ varies from $0$ to $150$.}\label{fig_sample_efficiency}
\end{figure}

Finally, compared with top-$k$ and random-$k$ for $k$ up to $150$, we find that our approach matches top-$k$ selection when $k = 150$ ($9000$ tokens) with $10$ in-context examples ($600$ tokens). See illustration in Figure~\ref{fig_sample_efficiency}. In this setting, our approach requires only $13$ GB of GPU memory, whereas top-$k$ and random-$k$ selection use $47$ GB to achieve similar performance, all of which are tested on the DeepSeek-7B model. Our method also performs well on other models, such as Llama-8B and Qwen-7B. On the three reasoning tasks, \acronymre{} outperforms the most competitive baseline by $22\%$ on average, while using $49\%$ less computation.

Lastly, we note that \estimation{} can adapt to any ensemble method in principle. For example, we can extend our approach to accelerate BRIDGE~\cite{wan2025from}, resulting in comparable performance while reducing the number of GPU hours by $80\%$.

\subsection{Ablation Studies}

Next, we describe the parameter choices of our algorithms. We also extend our approach to handle unlabeled demonstrations in Appendix~\ref{app_extensions}.

\textit{Random ensemble.} \acronymre{} involves three tunable hyperparameters: the number of subsets $m$, subset size $k$, and projection dimension $d$. We vary $m$ from 500 to 2000 and find that scores converge when $m > 2\times n^{\text{Demo}}$. We set $k$ between 3 and 8. $k$ is mostly $5$, depending on the effective number of demonstrations. For $d$, we vary it from $200$ to $1000$ and observe that values beyond $400$ yield minimal gains, so we fix $d = 400$.

\textit{Forward selection.} In \acronymfs{}, each iteration uses the previously selected subset as the anchor for gradient-based estimation. When the subset is small, approximation errors can be high. We vary the start step of estimation $t$. We find that setting $t > 3$ yields the same performance as running the forward selection fully.

\textit{Dimension reduction.} The memory overhead of our approach is similar to standard in-context inference, with extra storage from computing gradients on input embeddings. In practice, this adds less than $30\%$ storage over standard model inference.
To achieve this result, we apply random projection to reduce the dimension of the gradients, while provably preserving accuracy. In our experiments, we project the gradients from a dimension of $4,096$ times the sequence length (e.g., $128$) down to $400$, and find that the gradient estimation yields an error within $1\%$ to the model outputs.

\textit{Anchor selection.}
We evaluate random anchor selection on the SST-2 dataset. We vary the number of anchor prompts from $1$ to $10$ and find that using $5$ anchor prompts is sufficient to reduce the approximation error of gradient estimation below $1\%$.
We also consider mean embedding selection, in which anchor prompts are chosen based on their embeddings being closest to the mean of all candidate embeddings. The results differ from randomly selecting demonstrations by less than $2\%$.

\section{Related Work}

Language models have shown strong in-context learning (ICL) capabilities, where they can adapt to new tasks during inference by conditioning on a few input-label pairs without requiring parameter updates.
\citet{min2022rethinking} provide an empirical analysis of several factors affecting ICL, including label space, input distribution, and sequence format. The effectiveness of ICL has been partly attributed to parallel structures in pretraining corpora, where phrase pairs follow similar templates within the context window \citep{chen2024parallel}, which is crucial in pretraining \cite{gururangan2020don}).

Building on these insights, a growing line of work has focused on improving demonstration selection and usage. ADAPT \cite{ross2024toward} introduces a framework in which a teacher model diagnoses student misconceptions and adaptively selects demonstrations. \citet{yu2025dynamic} propose a reinforcement learning–based ranking method that updates retrieval rankings using feedback from the LLM, enabling better performance on long-tail samples. \citet{wan2025from} take a different perspective by fitting a Gaussian process regression model that maps subsets of demonstrations to model losses.
Another direction involves ensemble-based strategies that combine multiple subsets of demonstrations. \citet{khalifa2023exploring} conduct inference separately on multiple subsets and aggregate the predictions using weighted ensembles determined by input similarity. Similarly, \citet{huang2024divide} partition demonstrations into groups and reweight output logits through a non-gradient optimization procedure.

It is also worth noting earlier research on feature selection in unsupervised learning, such as methods based on expectation-maximization and clustering \cite{dy2004feature}.
Recent work has advanced the theoretical understanding of in-context learning (ICL). \citet{zhang2024trained} analyze transformers with a single linear self-attention layer and show that such models can in-context learn linear functions. 
\citet{wu2024many} establish complexity bounds for attaining Bayes-optimal ICL with single-layer linear attention models.
We refer readers to the recent survey by \citet{dong2022survey} for a more comprehensive review of in-context learning.

The use of gradients for selection and data influence analysis has been extensively explored in multitask learning. Recent theoretical work has modeled information transfer in multi-task neural networks \citep{wu2020understanding}, with precise characterizations established in the linear regression setting \citep{yang2025precise}. Influence functions \citep{koh2017understanding} quantify the effect of perturbing individual training examples on model predictions. PCGrad \citep{yu2020gradient} mitigates destructive gradient interference by projecting conflicting gradients in the shared parameter space.

Gradients have also been leveraged for attribution and task relationship modeling. \citet{fifty2021efficiently} propose task-affinity grouping, which constructs a task affinity matrix using gradient similarity across tasks. Building on this idea, \citet{li2023identification} introduce a surrogate modeling framework that extrapolates value functions on subset combinations to capture multi-task relationships. Follow-up work by \citet{li2023boosting} improves this framework with ensemble methods and more accurate task affinity computation, further extended by \citet{li2024scalable}. More recently, \citet{li2025efficient} apply this ensemble methodology to low-rank fine-tuning of LLMs.
Our work adds to this literature by employing gradients to approximate model \emph{inference}, enabling the development of a linear-time algorithm. We further anticipate that this methodology could scale to broader inference paradigms, including LLM alignment \citep{huang2024deal}.

\section{Conclusion}

We study the problem of demonstration selection for in-context learning.
Our key observation is that the first-order approximations of model outputs are highly accurate when the input embeddings are close.
Leveraging this, we propose an estimation algorithm that uses first-order approximations around an anchor prompt to efficiently estimate the model outputs for all other prompts. 
We validate our method on sentiment and reasoning tasks using various language models, achieving higher accuracy with less computation for in-context learning.

\section*{Acknowledgments}

Thanks to the anonymous referees and the action editor for their constructive feedback.
The work of Z. Zhang, Z. Zhang, D. Li, and H. R. Zhang is partially funded by NSF award IIS-2412008.
D. Li is also partially funded by a PhD fellowship from JPMorgan Chase \& Co.
Any views or opinions expressed herein are solely those of the authors listed, and may differ from those expressed by NSF or JPMorgan Chase.

\section*{Limitations and Future Works}

Our method selects in-context examples based on estimated inference results within the training distribution. However, its robustness in out-of-distribution (OOD) scenarios remains unclear.
Additionally, more principled anchor selection strategies, such as integrating with prior feature selection methods for unsupervised learning, can be explored in future work.
Our gradient-based approximation framework provides a flexible foundation for such extensions, including the acceleration of other prompt optimization methods.

\section*{Potential Risks}

This paper studies data selection for prompt-tuning language models. While language models may have future societal impacts, due to the technical focus of our work, we believe this paper raises minimal such concerns or negative implications.

%% file: appendix.tex
\section{Algorithms}

In this section, we elaborate on the algorithm description from Section \ref{sec_approach}.

\subsection{Subset Selection}\label{motivation_details}

In this section, we give several examples of conventional subset selection methods.

\begin{example}[Forward selection]\label{ex_fs}
    Forward selection iteratively selects a demonstration example into the subset that leads to the smallest loss at each iteration.
    It starts with an empty set $S_1 = \set{}$. Then, enumerate through all singleton sets, $f(\set{1}), \dots, f(\set{n})$, and pick the best one. Suppose $i_1$ is chosen, then iterate over $f(\set{1, i_1}), \ldots, f({n, i_1})$ except when $i_1$ is repeated and choose the best one. This procedure stops when $f(S)$ reaches its peak.
\end{example}

\smallskip
\begin{example}[Random ensemble]\label{ex_re}
     Random ensemble has been a highly effective strategy for data attribution~\citep{li2025efficient}, which can be used for subset selection.
     We draw $m$ random subsets from $\{1, 2, \dots, n\}$, each of fixed size $\alpha$. Then, evaluate $f(S)$ for each of the subsets and compute a score for a demonstration $z_i$ as the average of $f(S)$ over subsets $S$ that include $z_i$. The selection can be conducted by choosing the demonstrations with the lowest scores.
\end{example}

\smallskip
\begin{example}[Cross-entropy loss-based selection]\label{ex_loss}
    Existing work~\citep{peng2024revisiting} has designed a selection method that first reduces the search space by retrieving $K$ most similar demonstrations to the test examples. This is achieved by selecting $K$ demonstrations whose embeddings have the least distance to the embedding of the test query. Then, enumerate to evaluate $f(S)$ on each of $K$ demonstrations and select the $k$ demonstrations with the lowest losses.
\end{example}

When we construct the prompt condition within $\phi(S, x)$, we use a random ordering of the demonstrations in $S$, and we find that the ordering has a negligible effect on the predictions.
We use this convention throughout the paper.

\begin{algorithm}[h!]
\caption{In-Context Learning via Gradient-based Forward Selection (\acronymfs{})}\label{alg_forward_selection}
\textbf{Input:} Query set $\cS^{\text{Train}}$; Demonstration set $\cS^{\text{Demo}}$ \\
\textbf{Require:} Model $f_W$; Embedding function $\phi(\cdot)$; Projection dimension $d$; Subset size $\alpha$\\
\textbf{Output:} Selected subset $S_k \subseteq \cS^{\text{Demo}}$ of size $k$
\begin{algorithmic}[1]
    \State $S \leftarrow \emptyset$
    \While{$|S| < k$}
        \State $m \leftarrow n^{\text{Demo}} - \abs{S}$
        \State $\set{S_{i_j} = S \cup \set{i_j} \mid j = 1, \dots, m} \leftarrow$ For any $i_j$-th element of $\cS^{\text{Demo}}$ not in $S$
        \State $S_0 \leftarrow$ Randomly choose one anchor subset from $S_{i_1}, S_{i_2}, \dots, S_{i_m}$
        \State $\hat h(S_{i_1}), \hat h(S_{i_2}), \dots \hat h(S_{i_m}) \leftarrow$ \estimation$(S_0; \set{S_{i_j}}_{j=1}^m)$
        \State $i_{j^\star} \gets \arg\min_{j=1}^m \{\hat h(S_{i_j}) \}$
        \State $S \gets S_{i_{j^\star}}$
    \EndWhile\\
    \Return $S$
\end{algorithmic}
\end{algorithm}

\begin{algorithm}[h!]
\caption{In-Context Learning via Gradient Estimation of Cross Entropy (\acronymcone{})}\label{alg_cross_entropy}
\textbf{Input:} Query set $\cS^{\text{Train}}$; Demonstration set $\cS^{\text{Demo}}$ \\
\textbf{Require:} Model $f_W$; Embedding function $\phi(\cdot)$; Projection dimension $d$; Subset size $\alpha$\\
\textbf{Output:} Selected subset $S_k \subseteq \cS^{\text{Demo}}$ of size $k$
\begin{algorithmic}[1]
    \State $\cS^{(K)} \subseteq \cS^{\text{Demo}} \leftarrow$ The $K$ demonstrations most similar relative to Average$\big(\set{ \phi(x_i^{\text{Train}}) }\big)$ over the training set
    \State $S_0 \in \cS^{(K)} \leftarrow$ Randomly choose one sample
    \For {$i=1, 2, \dots, K$}
        \State $\set{\hat h(S_i)}_{i=1}^K \leftarrow$ \estimation$(S_0; S_i)$
    \EndFor
    \State $S_k \leftarrow$ The $k$ samples in $\cS^{(K)}$ that have the lowest values of $\set{\hat h(S_i)}_{i=1}^K$
    \State \Return $S_k$
\end{algorithmic}
\end{algorithm}

\subsection{Complete Procedure}\label{app_omitted_approximation_error}

We now describe instantiations of our approach for the above subset selection methods.
In particular, we provide the entire procedure for accelerating forward selection in Algorithm~\ref{alg_forward_selection} and loss-based selection in Algorithm~\ref{alg_cross_entropy}.

\begin{table*}[t!]
\centering
\caption{We report the RSS of the first-order approximation for the model inference output on the SST-2 dataset for various models of $1$ billion to $34$ billion parameters. As shown in the table, the relative error is the order of $10^{-4}$ to $10^{-2}$, where the relative distance of input embeddings is from $15\%$ to $40\%$. To obtain the distance range, we vary the value of $k$ from $2$ to $50$ and sample random combinations whose size falls within this range. We measure the deviation in $x$ and bucket the results. The MSE of the first-order approximation across different datasets and model scales during inference. Each entry reports the average approximation error and its standard deviation, measured with $k=4$ in-context examples. The models are listed in ascending order of parameter size to facilitate comparison across scales.}\label{table_approx_first_order}
{\small
\begin{tabular}{@{}lccccccc@{}}
\toprule
{Relative Distance} & {Llama-3.2-1B} & {OPT-1.3B} & {Llama-3.2-3B} & {Llama-3.1-8B} \\ 
\midrule
$15\%-20\%$~~~~~~~~~~~    & $1.6_{\pm0.3}\times10^{-3}$ & $4.8_{\pm2.2}\times10^{-3}$ & $8.4_{\pm1.3}\times10^{-3}$ & $6.8_{\pm1.5}\times10^{-3}$ \\
$20\%-25\%$~~~~~~~~~~~    & $2.1_{\pm0.1}\times10^{-3}$ & $5.7_{\pm0.8}\times10^{-3}$ & $9.2_{\pm0.9}\times10^{-3}$ & $1.9_{\pm0.4}\times10^{-2}$ \\
$25\%-30\%$~~~~~~~~~~~    & $2.5_{\pm0.7}\times10^{-3}$ & $8.6_{\pm0.4}\times10^{-3}$ & $1.0_{\pm0.2}\times10^{-2}$ & $2.4_{\pm0.6}\times10^{-2}$ \\
$30\%-35\%$~~~~~~~~~~~    & $6.9_{\pm0.3}\times10^{-3}$ & $1.1_{\pm0.1}\times10^{-2}$ & $1.2_{\pm0.1}\times10^{-2}$ & $4.6_{\pm0.7}\times10^{-2}$ \\
$35\%-40\%$~~~~~~~~~~~    & $9.6_{\pm0.4}\times10^{-3}$ & $1.6_{\pm0.2}\times10^{-2}$ & $2.8_{\pm0.3}\times10^{-2}$ & $5.2_{\pm1.7}\times10^{-2}$ \\
\bottomrule
\toprule
{Language Model} & {Poem-Sentiment} & {Edge-Existence}  & {CR} & {Modular-Addition} \\ 
\midrule
{Llama-3.2-1B} & $3.7_{\pm0.1} \times 10^{-3}$ & $8.9_{\pm0.7} \times 10^{-4}$& $1.8_{\pm0.3} \times 10^{-2}$& $6.1_{\pm1.6} \times 10^{-3}$\\
{OPT-1.3B} & $1.4_{\pm0.3} \times 10^{-2}$ & $1.1_{\pm0.1} \times 10^{-2}$ & $2.5_{\pm0.6} \times 10^{-2}$ & $2.3_{\pm1.1} \times 10^{-3}$\\
{Llama-3.2-3B} & $1.7_{\pm0.7} \times 10^{-3}$ & $6.1_{\pm0.2} \times 10^{-3}$ & $3.3_{\pm1.0} \times 10^{-3}$ & $3.5_{\pm1.0} \times 10^{-3}$\\
{DeepSeek-7B}  & $2.8_{\pm0.9} \times 10^{-3}$ & $1.6_{\pm0.2} \times 10^{-3}$& $1.7_{\pm0.9} \times 10^{-2}$& $3.2_{\pm0.1} \times 10^{-3}$\\
{Llama-3.1-8B} & $4.3_{\pm0.6} \times 10^{-3}$ & $1.6_{\pm0.1} \times 10^{-2}$ & $1.8_{\pm0.1} \times 10^{-2}$ & $4.2_{\pm1.2} \times 10^{-3}$\\
{Llama-2-13B}  & $1.0_{\pm0.6} \times 10^{-3}$ & $1.4_{\pm0.0} \times 10^{-3}$ & $1.6_{\pm0.4} \times 10^{-4}$ & $8.3_{\pm1.6} \times 10^{-3}$\\
{CodeLlama-34B} & $4.9_{\pm2.9} \times 10^{-4}$ & $6.4_{\pm1.0} \times 10^{-4}$  & $1.0_{\pm0.2} \times 10^{-3}$ & $1.6_{\pm0.2} \times 10^{-3}$\\
\bottomrule
\end{tabular}}
\end{table*}

\acronymfs{} adopts a forward selection strategy, where at each step the method greedily selects the next demonstration that leads to the lowest estimated inference loss on the validation input. The selection continues until $k$ demonstrations are chosen. This procedure aims to construct an effective prompt by iteratively adding the most helpful examples based on the current context.

\acronymre{} generates a large number of random $k$-way demonstration combinations, then uses inference to compute the loss of each combination. Each sample is scored by averaging the losses of all combinations that include it. Finally, the $k$ samples with the lowest average scores are selected. This ensemble-based approach provides a robust estimation by aggregating performance across many contexts.

We use the Johnson–Lindenstrauss Lemma to perform dimensionality reduction, with the detailed formulation and implementation provided in Appendix~\ref{app_JL}. After projection, the dimensionality grows only logarithmically with the number of data points and decreases with the square of the allowable distortion. This lemma ensures that the Euclidean distance between any two vectors is approximately preserved with high probability.

As a remark, our method estimates model outputs through a first-order approximation in the input embedding space. This idea can be further extended to other inference procedures. Since the gradient depends only on the reference prompt, one may calibrate the estimation by adjusting the combination weights at different positions. In particular, the weights can be modulated according to the inner product between the gradient and the embedding differences, enabling position-specific adjustments. Moreover, our method naturally applies to the prompt engineering setting, where only a single token changes in each inference. The development of such extensions is left for future work.

\begin{table*}[t!]
\centering
\caption{We report the relative residual sum of squares between $f(S)$ and $\hat{f}(S)$ when running the subset selection baselines, measured on the Coin-Flip dataset. To measure speedup, we report the speedup rate as the ratio of FLOPs required to fully compute $f(S)$ versus \acronym{}. The speedup remains consistent across different models on the same dataset, as the reduced number of forward passes is constant for all models on the same dataset.}\label{table_approx_error}
{\small\begin{tabular}{@{}lcccccc|c@{}}
\toprule
    Coin-Flip & {{Llama-3-1B}} & {{Llama-3-3B}} & {{Llama-3-8B}} & {{Llama-2-13B}} & {Speedup} \\ \midrule
    \acronymfs{} & $0.15_{\pm0.01} \%$  & $0.51_{\pm0.03} \%$  & $0.88_{\pm0.00} \%$  & $0.62_{\pm0.01} \%$ & $\mathbf{37.7\times}$ \\
    \acronymre{} & $0.17_{\pm0.01} \%$ & $0.15_{\pm0.01} \%$ & $0.37_{\pm0.02} \%$ & $0.21_{\pm0.00} \%$ & $\mathbf{19.7\times}$ \\
    \acronymcone{} & $0.24_{\pm0.00} \%$ & $0.25_{\pm0.01} \%$ & $0.90_{\pm0.10} \%$ & $0.68_{\pm0.01} \%$ & $\mathbf{17.3\times}$ \\ \midrule
    SST-2 & {{Llama-3-1B}} & {{Llama-3-3B}} & {{Llama-3-8B}} & {{Llama-2-13B}} & {Speedup} \\ \midrule
    \acronymfs{}   & $ 0.12_{\pm0.07} \%$ & $ 0.45_{\pm0.27} \%$ & $ 0.39_{\pm0.04} \%$ & $ 1.50_{\pm0.40} \%$ & $\mathbf{19.0\times}$ \\
    \acronymre{}   & $ 0.50_{\pm0.17} \%$ & $ 0.42_{\pm0.12} \%$ & $ 0.50_{\pm0.19} \%$ & $ 0.40_{\pm0.19} \%$ & $\mathbf{10.8\times}$ \\
    \acronymcone{} & $ 0.93_{\pm0.11} \%$ & $ 0.88_{\pm0.12} \%$ & $ 1.30_{\pm0.40} \%$ & $ 0.46_{\pm0.04} \%$ & $\mathbf{6.7\times}$ \\    
    \bottomrule
\end{tabular}}
\end{table*}

\subsection{Random Projection}\label{app_JL}

We utilize the Johnson-Lindenstrauss Lemma to reduce the storage of gradients. Here we describe the details of it. This result is used to project vectors from high-dimensional Euclidean space into a low-dimensional space, preserving distances between points nearly perfectly.

\begin{lemma}[The Johnson-Lindenstrauss Lemma]\label{lemma_jl}
Let $0 < \varepsilon < 1$, let $X$ be a set of $N$ points in $\mathbb{R}^n$, and let 
$k \;\geq\; C \,\varepsilon^{-2} \log N,$
for some universal constant $C > 0$.
Then there exists a linear map $f : \mathbb{R}^n \to \mathbb{R}^k$ such that for all $u,v \in X$,
\[ \left| \|f(u)-f(v)\|^2 - \|u-v\|^2 \right| \le \varepsilon\|u-v\|^2. \]
\end{lemma}

Equivalently, the restriction $f|_X$ is $(1+\varepsilon)$-bi-Lipschitz.  
Moreover, the bound on $k$ is asymptotically tight: there exist sets of $N$ points that require 
$k = \Omega\!\left(\frac{\log N}{\varepsilon^2}\right)$,
dimensions to preserve all pairwise distances within a factor of $(1\pm \varepsilon)$.

\smallskip\textit{Dimension reduction of gradients.}
Suppose we have $N$ gradient vectors $\{g_t\}_{t=1}^N \subset \mathbb{R}^d$.  
To compress them, we generate a random projection matrix 
$\Pi \in \mathbb{R}^{k\times d}$, where every $\Pi_{ij}$ is drawn independently from $\mathcal{N}(0,1)/\sqrt{k}$.
That is, each entry is sampled independently from a Gaussian distribution and scaled by $1/\sqrt{k}$.  
For each gradient $g_t$, we compute its compressed representation
$\tilde{g}_t = \Pi g_t \in \mathbb{R}^k.$
By Lemma \ref{lemma_jl}, with high probability, all pairwise distances and inner products among $\{g_t\}$ are preserved up to $(1\pm \varepsilon)$ after projection.  
Thus, storing $\tilde{g}_t$ instead of $g_t$ reduces the memory cost from $O(d)$ per gradient to $O(k)$ while maintaining the geometric relations needed for downstream analysis.

\smallskip\textit{Gradient estimation results.}
To further validate the robustness of our first-order loss approximation method, we report the approximation error, measured as the mean squared error (MSE) between estimated and actual inference losses. We evaluate this across six datasets using a variety of open-source language models ranging in size from $1$ billion to $34$ billion parameters.
In particular, we consider different datasets here, but we fix $k = 4$ instead.
The results, shown in Table \ref{table_approx_error}, remain consistent with our findings.
Overall, the error remains consistently low, within the range of $10^{-3}$ to $10^{-2}$ across most tasks. Notably, there is no clear upward trend in approximation error with increasing model size. For example, CodeLLaMA-34B, the largest model in our evaluation, still achieves errors comparable to or even smaller than those of smaller models on datasets such as GraphQA and SST-2. Since we do the approximation on the embedding space, the error may not strictly follow scaling laws. This observation aligns with the trend reported in Table~\ref{table_motivation_result}, suggesting that larger models do not affect the quality of our approximation.

\begin{table*}[t!]
\centering
\caption{Total number of floating-point operations (FLOPs) required during inference for different datasets using the DeepSeek-LLM-7B and LLaMA-3.1-8B models. We report the results on six datasets with varied task categories.}\label{table_computation_cost}
\resizebox{1.00\textwidth}{!}
{\small
\begin{tabular}{@{}lcccccc@{}}
\toprule
DeepSeek-7B & {Poem Sentiment} & {SST-2} & {CR} & {Edge Existence} & {Modular Addition} & {Coin Flip}\\
\midrule
Forward Selection & $1.1 \times 10^{16}$ & $2.8 \times 10^{15}$& $2.8 \times 10^{15}$& $3.0 \times 10^{16}$ & $8.7 \times 10^{15}$& $2.5 \times 10^{16}$\\
Top-$k$& $2.1 \times 10^{14}$ & $1.1 \times 10^{14}$ & $1.1 \times 10^{14}$ & $5.7 \times 10^{14}$ & $3.3 \times 10^{14}$ & $2.8 \times 10^{14}$ \\
Top-$k$ + CE & $2.9 \times 10^{15}$ & $7.8 \times 10^{14}$ & $7.9 \times 10^{14}$ & $8.0 \times 10^{15}$ & $2.4 \times 10^{15}$ & $6.5 \times 10^{15}$ \\
UR & $2.0 \times 10^{15}$ & $5.1 \times 10^{14}$ & $5.2 \times 10^{14}$ & $5.7 \times 10^{15}$ & $1.6 \times 10^{15}$ & $4.8 \times 10^{15}$ \\
\acronymcone{}& $2.5 \times 10^{14}$ & $1.2 \times 10^{14}$ & $1.2 \times 10^{14}$ & $6.9 \times 10^{14}$ & $3.6 \times 10^{14}$ & $3.8 \times 10^{14}$ \\
\acronymre{}& $\bm{4.2 \times 10^{13}}$ & $\bm{1.1 \times 10^{13}}$ & $\bm{1.1 \times 10^{13}}$ & $\bm{1.2 \times 10^{14}}$ & $\bm{3.3 \times 10^{13}}$ & $\bm{9.8 \times 10^{13}}$ \\
\acronymfs{}& $3.8 \times 10^{14}$ & $1.5 \times 10^{14}$ & $1.5 \times 10^{14}$ & $1.0 \times 10^{15}$ & $4.6 \times 10^{14}$ & $6.7 \times 10^{14}$ \\
\midrule
Llama-3.1-8B & {Poem Sentiment} & {SST-2} & {CR} & {Edge Existence} & {Modular Addition} & {Coin Flip}\\
\midrule
Forward Selection& $1.8 \times 10^{16}$& $4.5 \times 10^{16}$& $4.6 \times 10^{16}$& $4.8 \times 10^{17}$& $1.4 \times 10^{17}$& $4.0 \times 10^{17}$\\
Top-$k$& $3.4 \times 10^{14}$& $1.7 \times 10^{14}$& $1.7 \times 10^{14}$& $9.2 \times 10^{14}$& $5.3 \times 10^{14}$& $4.5 \times 10^{14}$\\
Top-$k$ + CE & $4.6 \times 10^{15}$& $1.2 \times 10^{15}$& $1.3 \times 10^{15}$& $1.3 \times 10^{16}$& $3.9 \times 10^{15}$& $1.0 \times 10^{16}$\\
UR & $3.3 \times 10^{15}$& $8.2 \times 10^{14}$& $8.4 \times 10^{14}$& $9.1 \times 10^{15}$& $2.6 \times 10^{15}$& $7.6 \times 10^{15}$\\
\acronymcone{}& $4.0 \times 10^{14}$& $1.9 \times 10^{14}$& $1.9 \times 10^{14}$& $1.1 \times 10^{15}$& $5.8 \times 10^{14}$& $6.0 \times 10^{14}$\\
\acronymre{}& $\bm{6.7 \times 10^{13}}$& $\bm{1.7 \times 10^{13}}$& $\bm{1.8 \times 10^{13}}$& $\bm{1.9 \times 10^{14}}$& $\bm{5.4 \times 10^{13}}$& $\bm{1.6 \times 10^{14}}$\\
\acronymfs{}& $6.0 \times 10^{14}$& $2.4 \times 10^{14}$& $2.4 \times 10^{14}$& $1.7 \times 10^{15}$& $7.4 \times 10^{14}$& $1.1 \times 10^{15}$\\
\bottomrule
\end{tabular}}
\end{table*}

\section{Omitted Experiments}\label{sec_appendix_exp}

In this section, we present additional details regarding the experiment setup, report supporting results and analysis, and describe extensions of our approach.

\subsection{Detailed Setup}\label{appendix:datasets_models}

\textit{Datasets.} Our datasets cover a diverse set of task types, including linguistic tasks, algorithmic reasoning, arithmetic reasoning, and graph reasoning.

The \href{https://huggingface.co/datasets/google-research-datasets/poem_sentiment}{Poem Sentiment} dataset contains sentiment annotations (positive, neutral, or negative) for lines of poetry, supporting sentiment analysis in literary contexts. The number of queries is $800$.

The \href{https://huggingface.co/datasets/nyu-mll/glue/viewer/sst2}{SST-2} dataset is a binary sentiment classification dataset composed of movie reviews labeled as either positive or negative from the GLUE benchmark. The number of queries is $450$.

The \href{https://huggingface.co/datasets/SetFit/CR}{Customer Review (CR)} dataset includes product reviews annotated with binary sentiment labels, designed to assess the model’s ability to generalize sentiment understanding across domains. The number of queries is $450$.

Modular addition is a math reasoning task where models take $a,b \in \{0,..., P-1\}$ for a prime $P$ and predict their sum mod $P$. The number of queries is 500.

The Edge Existence dataset from \href{https://github.com/google-research/google-research/tree/master/graphqa}{GraphQA} is a graph reasoning task in which the model is given an undirected graph and must determine whether a specific edge is present in this graph. The number of queries is $799$.

The \href{https://huggingface.co/datasets/skrishna/coin_flip}{Coin-Flip} dataset is an arithmetic reasoning task where the model reads a natural language description of a sequence of fair coin flips and must predict the final outcome (heads or tails). The number of queries is $869$.

\smallskip\textit{Models.} In our experiments, we evaluate a range of open-source language models with varying scales and architectures. These include \href{https://huggingface.co/meta-llama/Llama-3.2-1B}{LLaMA-3.2-1B}, \href{https://huggingface.co/facebook/opt-1.3b}{OPT-1.3B}, \href{https://huggingface.co/meta-llama/Llama-3.2-3B}{LLaMA-3.2-3B}, \href{https://huggingface.co/deepseek-ai/deepseek-llm-7b-chat}{DeepSeek-LLM-7B}, 
\href{https://huggingface.co/Qwen/Qwen2.5-7B-Instruct}{Qwen2.5-7B-Instruct},
\href{https://huggingface.co/Qwen/Qwen3-8B}{Qwen3-8B},
\href{https://huggingface.co/meta-llama/Llama-3.1-8B}{LLaMA-3.1-8B}, \href{https://huggingface.co/meta-llama/Llama-2-13b-hf}{LLaMA-2-13B}, and \href{https://huggingface.co/codellama/CodeLlama-34b-hf}{CodeLLaMA-34B}.
This diverse selection enables us to evaluate the performance and scalability of our methods across models of varying capacities, training frameworks, and intended usage scenarios.

\smallskip\textit{Baselines.} We compare our algorithm to baselines including random-$k$, BM25, top-$k$, top-$k$ + CE, and UR. All these selection methods rely on input features, which are extracted from the encoder output of the models.

The random-$k$ \citep{min2022rethinking} method randomly selects $k$ samples from the candidate set.
The top-$k$ \citep{liu2022makes} method selects the $k$ most similar candidates based on feature similarity, given a test input. We compute the cosine similarity in the top-$k$ range between the last-layer hidden representations of training queries and demonstration examples, based on a specific model.
The BM25 method utilizes a term-frequency-based ranking function to retrieve the top-$k$ candidates whose input texts are most relevant to the query, with a focus on lexical overlap rather than embedding-level similarity.
These three baselines only make the selection based on the input features, without considering the inference result, which may not accurately measure sample similarity. For example, in the reasoning datasets, such as the coin flip, the sample can vary by one token in the input but have different labels.
The top-$k$ + CE \citep{peng2024revisiting} method follows a two-step process. The first step is to filter $k$ samples, and the second step selects the sample that results in the smallest output loss during inference. This structured approach ensures that the selection methods effectively leverage feature representations for candidate retrieval. 
UR \citep{yu2025dynamic} applies BM25 to pre-select a candidate pool for each query. Then, in the training phase, it repeatedly runs the language model with 0 to $k$ retrieved examples, and assigns reward scores to each example based on whether adding it improves or harms the model prediction. In the inference phase, it again uses BM25 to filter candidates and ranks them based on the reward scores learned during training, ultimately selecting the top-$k$ highest-scoring demonstration examples. BRIDGE~\cite{wan2025from} proposes selecting demonstration subsets by fitting a Gaussian process regression model to the output of the demonstration subsets.

\subsection{Additional Experiment Results}\label{app_computation}

In Table \ref{table_computation_cost}, we report the number of FLOPs for each approach, corresponding to Table \ref{table_inference_error}.

To evaluate the performance on reasoning tasks, we further assess our approach on two reasoning tasks: Snarks and Sports Understanding from the BIG-Bench-Hard benchmark, and report the $F_1$ score in Table~\ref{tab_big_bench}. We find that the gradient-based estimation yields a relative error of less than $1\%$ when the relative distance of the input embedding is within $10\%$. When applied to demonstration selection, our approach also leads to a $18.5\%$ performance increase compared to top-$k$.
The results show that applying our approach yields comparable performance to BRIDGE while using only $20\%$ of GPU hours.

\begin{table}[t!]
\centering
\caption{We report the result using Snarks and Sports Understanding dataset.}\label{tab_big_bench}
{\small\begin{tabular}{@{}lcc@{}}
\toprule
Algorithm & Snarks & Sports Understanding \\ \midrule
Top-$k$ & $0.33$ & $0.35$\\
BRIDGE & $0.50$ & $0.48$\\
\acronymre{} & $\bm{0.53}$ & $\bm{0.52}$\\
\bottomrule
\end{tabular}}
\end{table}

We also evaluate our approach with another instruction-tuned model, Qwen-7B-Instruct. First, we find that on this model, the gradient-based estimation still yields relative error below $1\%$ within a $30\%$ relative distance in the input embeddings space. Additionally, we evaluate demonstration selection using the models on SST-2 and modular addition datasets. We report the results in Table~\ref{tab_qwen}.

\begin{table}[h!]
\centering
\caption{We report the result of \acronymre{} on SST-2 and modular addition using Qwen-7B-Instruct model.}\label{tab_qwen}
{\small\begin{tabular}{@{}lcc@{}}
\toprule
Datasets & SST-2 & Modular Addition \\ \midrule
Random-$k$ & $0.87$ & $0.68$\\
Top-$k$ & $0.72$ & $0.61$\\
\acronymre{} & $\bm{0.89}$ & $\bm{0.73}$\\
\bottomrule
\end{tabular}}
\end{table}

\subsection{Additional Ablation Analysis}\label{app_ablation}

\noindent\textit{Number of anchor subsets in random ensemble.}
Recall that \acronymre{} pre-computes model inference losses on a few anchor subsets and estimates losses for other prompts via first-order approximation. On the Edge Existence dataset, we vary the number of anchor prompts from $1$ to $10$.
We find that $5$ anchor prompts are sufficient to keep the estimation error below $5\%$, beyond which there are minimal gains for in-context learning performance. Thus, we set the number of anchors as $5$.

\smallskip\textit{Duplicated samples.} When the candidate sample set contains duplicate samples, it can affect the selection results of similarity-based methods, as these methods consider only the input data itself, assigning identical similarity scores to duplicate samples. To investigate the impact of repeated samples, we conduct experiments on the StrategyQA dataset by duplicating each candidate sample three times (Table~\ref{tab_duplicate}), allowing us to analyze how redundancy influences the selection process and overall model performance.
\begin{table}[t!]
\centering
\caption{We illustrate the results on the StrategyQA dataset when the candidate data is duplicated three times. For the top-$k$ algorithm, a small number of repeated samples ($k=2$) can improve performance, but additional repetition ($k=3$) reduces performance. After the model has selected all repeated samples, choosing a new, non-duplicate sample ($k=4$) can once again improve performance, even surpassing the performance of the duplicated case when $k=2$. On the other hand, \acronym{} provides a better and stable performance when $k\geq2$.}\label{tab_duplicate}
{\small\begin{tabular}{@{}ccc@{}}
    \toprule
    \# Duplicates & {Top-$k$} & \textbf{\acronym{}} \\ \midrule
    $k=1$        & $31.51$ & $31.97$     \\
    $k=2$        & $38.01$ & $49.00$     \\
    $k=3$        & $32.52$ & $48.95$     \\
    $k=4$        & $39.29$ & $48.45$     \\ \bottomrule
\end{tabular}}
\end{table}

\smallskip\textit{Model size.} We use \acronymre{} to compare the in-context learning performance across different LMs, including LLaMA-1B, OPT-1.3B, LLaMA-3B, DeepSeek-7B, and LLaMA-8B.
We evaluate \acronymre{} on the SST-2 dataset, varying the number of demonstrations from $3$ to $8$.
We find that under the same number of demonstration examples, DeepSeek-7B performs the best across the six models. Additionally, we find that models with more parameters tend to achieve better in-context learning performances.

\smallskip\textit{Robustness of random ensemble.}
One of the main contributions of this paper is the development of an influence score for demonstration examples in in-context learning, utilizing a random ensemble approach (see Algorithm \ref{alg_random_ensemble}). This approach has been used to estimate data influence in prior work on data attribution and task selection (e.g., higher-order task affinity scores~\cite{fifty2021efficiently}), where it has been shown to capture higher-order relationships between samples or tasks. %
Our paper shows that the averaged random ensemble scores remain effective for selecting demonstrations for in-context learning. This is illustrated through a case study on linear functions, with extensions to nonlinear functions such as multi-layer perceptrons. We find that as the number of sampled subsets increases, the scores converge, allowing them to separate demonstrations drawn from the same function class from those from another function class.

We now elaborate on the example of linear functions~\cite{garg2022can}. As the number of sampled subsets increases, this results in a separation in the scores between good and poor examples. Suppose demonstrations are drawn from two linear functions with coefficients $\beta^{(1)}$ and $\beta^{(2)}$, and training queries are sampled from $\beta^{(1)}$. For subsets of demonstrations with more than 20 samples from $\beta^{(1)}$, the model yields near-zero loss. As we sample over 50 subsets, we observe a separation between the scores of demonstrations from $\beta^{(1)}$ and $\beta^{(2)}$, which indicates the effectiveness of random ensemble. We report the results in Table~\ref{tab_num_subsets}.

Note that in Table~\ref{table_motivation_result}, we have evaluated the average error across different ranges of distances between a random subset and the anchor prompt. Indeed, we noticed that when the distance is lower, the error tends to be smaller. With a relative distance of up to $40\%$, the error remains within $2\%$. In practice, we find that over $80\%$ of random subsets remain within $30\%$ distance to the anchor prompts.

\begin{table}[t!]
\centering
\caption{To illustrate the scoring mechanism in our random ensemble method, we report the average score of demonstration samples from $\beta^{(1)}$ and $\beta^{(2)}$, given the query set generated from $\beta^{(1)}$. We vary the number of subsets and find that the in-distribution samples always achieve lower loss than out-of-distribution samples.}\label{tab_num_subsets}
\resizebox{1.00\columnwidth}{!}
{\small\begin{tabular}{@{}lcccc@{}}
\toprule
\# Subsets & $5$ & $50$ & $100$ & $1000$ \\ 
\midrule
Samples from $\beta^{(1)}$ & $2.1_{\pm2.0}$ & $3.1_{\pm1.1}$ & $3.9_{\pm0.8}$ & $3.7_{\pm0.4}$  \\
Samples from $\beta^{(2)}$ & $2.6_{\pm3.6}$ & $4.6_{\pm2.7}$ & $5.2_{\pm2.5}$ & $5.4_{\pm2.7}$  \\
\bottomrule
\end{tabular}}
\end{table}

Additionally, we find that the loss of using different orders of demonstration examples is similar. On the SST-2 and modular addition dataset, we measure the variance in loss across different permutations of demonstration order. The number of demonstration examples is $8$. We find that the variance is less than $2\%$ of the mean loss, suggesting that the model is largely insensitive to the order of demonstrations. Based on this observation, we do not model order sensitivity in our approach.

\subsection{Extensions}\label{app_extensions}

\textit{Selection for individual test queries.}
Our approach can be directly applied to selecting demonstration examples for individual test queries by specifying a training query set for each test query. One method is to select training queries with feature representations that are closest to the given test query. We evaluate this on the SST-2 and Modular Addition datasets. We find that this method improves test accuracy by an average of $6\%$ compared to selecting a single demonstration subset.

We note that, in both settings of selecting demonstrations for each or all test queries, the computational cost of our approach is roughly the same. This is because our approach requires only computing the loss and gradients of anchor prompts on the training queries once, and then estimating the loss on other subsets using the gradient-based estimation.

\smallskip\textit{Unlabeled demonstrations.} The methods described above assume access to a labeled demonstration set. We can extend them to the unlabeled setting by generating pseudo-labels for the demonstration set using a two-step procedure: first, apply top-$k$ to select a labeled set of demonstrations $\hat{S}$. Second, use the prompt constructed from $\hat{S}$ to label the demonstration inputs. Then, we can apply our methods with the pseudo-labels. In our ablation studies, we find that pseudo-labels yield performance comparable to ground-truth labels, consistent with prior work \cite{min2022rethinking}.

To investigate the above, we evaluate \acronym{} under three labeling settings for the demonstration examples: (1) \textit{Pseudo-labels}, which are obtained from model predictions, (2) \textit{True labels}, which are the real output of the queries, and (3) \textit{Random labels}, where the label assignments are randomly shuffled. We conduct this analysis on the SST-2 and CR datasets using the DeepSeek-7B model, with $k=3$ prompt samples.

Remarkably, we observe that the choice of labels yields the same final performance, whether accurate, pseudo, or random labels. All three settings yield identical accuracy scores for both datasets.
For SST-2, we observe approximately $0.895$ for pseudo, true, or random labels, and around $0.742$ for CR, using the DeepSeek-LLM-7B model.

These findings suggest that the effectiveness of \acronym{} does not rely heavily on the semantic correctness of the labels within the selected demonstrations. Instead, the input content and structural patterns within the prompts play a dominant role in guiding the model's prediction. This observation aligns with the findings of \citet{min2022rethinking}, who showed that in-context learning can still perform well even when demonstrations contain incorrect or misleading label information.

\subsection{Model Generalization}

Finally, we assess model generalization by measuring sharpness.
We design an inference method by injecting random noise to compute the Hessian trace.
Specifically, we sample a $d$-dimensional isotropic Gaussian variable $U\sim\cN(0, \sigma^2\id)$.
Using Taylor's expansion, we have:
\begin{align*}
    f(x+U) =& f(x) + \inner{U}{\nabla f(x)} \\
    &+ \frac{1}{2} {U^{\top}}{[\nabla^2 f(x)]} U + O(\bignorms{U}^3).
\end{align*}
With $\ex{U} = 0$ and $\ex{UU^{\top}} = \sigma^2\id$, we get
\begin{align*}
    \exarg{U\sim\cP}{f(x + U)} \approx f(x) &+ \frac 1 2\sigma^2\tr[\nabla^2 f(x)],
\end{align*}
plus an error term that scales with $O(\sigma^3)$.
Suppose that $\sigma$ is a small value, we can get the following approximation of the Hessian trace:
\begin{align*}
    \tr[\nabla^2 f(x)] \approx \frac{2}{\sigma^2} \left( \exarg{U\sim\cP}{f(x + U)} - f(x) \right).
\end{align*}
We now examine the generalization ability of different prompt selection strategies. We define generalization as the extent to which a prompt selected for one query remains effective across other queries. Prior methods, such as top-$k$, which rely on embedding similarity, typically construct a separate prompt for each query in isolation. In contrast, \acronym{} selects demonstrations jointly across the entire evaluation set, enabling shared prompts.
\begin{table}[t!]
\centering
\caption{We measure the Hessian trace on the training queries and testing queries for in-context learning of linear functions. \acronym{} not only achieves the lowest loss, but also reduces the Hessian trace.
We set $k=25$ and run each experiment with $30$ random seeds to report the standard deviations.}\label{tab_lr_hessian}
{\small\begin{tabular}{lcc}
\toprule
Algorithms & Hessian (Training) & Hessian (Test) \\
\midrule
Random-$k$ & $7.85_{\pm 9.28}$ & $6.73_{\pm 2.47}$ \\
Top-$k$ & $5.84_{\pm6.16}$ & $5.78_{\pm 3.94}$ \\
\acronym{} & $\bm{4.15}_{\pm 2.06}$ & $\bm{4.94}_{\pm1.55}$ \\
\midrule
\midrule
 & Loss (Training) & Loss (Test) \\ 
\midrule
Random-$k$ & $0.127_{\pm 0.312}$ & $0.127_{\pm0.227}$ \\
Top-$k$ & $0.205_{\pm 0.378}$ & $0.066_{\pm0.101}$\\
\acronym{} & $\bm{0.001}_{\pm 0.002}$ & $\bm{0.003}_{\pm0.003}$ \\
\bottomrule
\end{tabular}}
\end{table}

The results, shown in Table \ref{tab_lr_hessian}, demonstrate that \acronym{} not only achieves the lowest loss across all values of $k$, but also consistently reduces the Hessian trace, indicating a smoother loss surface. This suggests that the prompt selected by \acronym{} is more robust and generalizes better across queries. In contrast, top-$k$ suffers from the highest loss, highlighting the limitation of per-query similarity-based selection.
It would be interesting to further explore the role of Hessians in understanding in-context learning \cite{ju2022robust,zhangnoise}.
For example, recent work has developed generalization bounds on the pretraining objective of large language models \cite{finzicompute}, which connects to the Hessian spectrum.